\documentclass[letterpaper]{article} % DO NOT CHANGE THIS
\usepackage{aaai25}  
\usepackage{times}  % DO NOT CHANGE THIS
\usepackage{helvet}  % DO NOT CHANGE THIS
\usepackage{courier}  % DO NOT CHANGE THIS
\usepackage[hyphens]{url}  % DO NOT CHANGE THIS
\usepackage{graphicx} % DO NOT CHANGE THIS
\usepackage{latexsym}
\usepackage{amssymb}
\usepackage{amsmath}
\usepackage{amsthm}
\usepackage{booktabs}
\usepackage{enumitem}
\usepackage{graphicx}
\usepackage{color}
\usepackage{bm}
\usepackage{multirow}
\usepackage{float}
\usepackage{subfig}
\usepackage[table]{xcolor}
\usepackage{tabularx}
\usepackage{subcaption}
\usepackage{natbib}  % DO NOT CHANGE THIS AND DO NOT ADD ANY OPTIONS TO IT
\usepackage{caption} % DO NOT CHANGE THIS AND DO NOT ADD ANY OPTIONS TO IT
\usepackage{algorithm}
\usepackage{algorithmic}

\usepackage{newfloat}
\usepackage{listings}
\DeclareCaptionStyle{ruled}{labelfont=normalfont,labelsep=colon,strut=off} % DO NOT CHANGE THIS
\floatstyle{ruled}
\newfloat{listing}{tb}{lst}{}
\floatname{listing}{Listing}
\nocopyright
\title{PLPP: Prompt Learning with Perplexity Is Self-Distillation for Vision-Language Models}
\author{
       Biao Liu\textsuperscript{\rm 1},
    Wenyi Fang\textsuperscript{\rm 2},
    Xiaoyu Wu\textsuperscript{\rm 2},
    Yang Zheng\textsuperscript{\rm 2},
    Zheng Hu\textsuperscript{\rm 2},
    Bo Yuan\textsuperscript{\rm 1}
}
\affiliations{
    \textsuperscript{\rm 1}Department of Computer Science and Engineering,
    Southern University of Science and Technology,
    Shenzhen,
    518055\\
    \textsuperscript{\rm 2}RAMS Lab,
    Huawei Technologies Co., Ltd.,
    Shenzhen, 518129
}

\usepackage{bibentry}
\begin{document}

\maketitle

\begin{abstract}
% Pre-trained Vision-Language (VL) models such as CLIP have demonstrated their excellent performance across numerous downstream tasks. A recent method, Context Optimization (CoOp), further improves the performance of VL models on downstream tasks by introducing prompt learning. CoOp optimizes a set of learnable vectors, aka prompt, and freezes the whole CLIP model. Nonetheless, only using CLIP loss to tune the prompts tends to cause models for overfitting on datasets for downstream tasks. To address this issue, our work proposes a plug-in prompt-regularization method called PLPP (Prompt Learning with PerPlexity). PLPP designs a two-step operation to compute the perplexity for prompts: (a) calculating cosine similarity between the weight of the embedding layer and prompts to get labels, (b) introducing a language model (LM) head that requires no training behind text encoder to output word probability distribution. PLPP can be integrated in any prompt learning methods during training. Meanwhile, we unveil that the essence of perplexity is inherently a form of self-distillation learning.The experiments conducted on four classification tasks indicate that PLPP exhibits superior performance compared to existing methods.
Pre-trained Vision-Language (VL) models such as CLIP have demonstrated their excellent performance across numerous downstream tasks. A recent method, Context Optimization (CoOp), further improves the performance of VL models on downstream tasks by introducing prompt learning. CoOp optimizes a set of learnable vectors, aka prompt, and freezes the whole CLIP model. However, relying solely on CLIP loss to fine-tune prompts can lead to models that are prone to overfitting on downstream task. To address this issue, we propose a plug-in prompt-regularization method called PLPP (Prompt Learning with PerPlexity), which use perplexity loss to regularize prompt learning. PLPP designs a two-step operation to compute the perplexity for prompts: (a) calculating cosine similarity between the weight of the embedding layer and prompts to get labels, (b) introducing a language model (LM) head that requires no training behind text encoder to output word probability distribution. Meanwhile, we unveil that the essence of PLPP is inherently a form of self-distillation. To further prevent overfitting as well as to reduce the additional computation introduced by PLPP, we turn the hard label to soft label and choose top-$k$ values for calculating the perplexity loss. For accelerating model convergence, we introduce mutual self-distillation learning, that is perplexity and inverted perplexity loss.  The experiments conducted on four classification tasks indicate that PLPP exhibits superior performance compared to existing methods.
\end{abstract}

% Uncomment the following to link to your code, datasets, an extended version or similar.
%
% \begin{links}
%     \link{Code}{https://aaai.org/example/code}
%     \link{Datasets}{https://aaai.org/example/datasets}
%     \link{Extended version}{https://aaai.org/example/extended-version}
% \end{links}
% \begin{figure}[tb]
% \centering
% %\includegraphics[width=3in]{fig5}
% \subfloat[CoOp]{
% 	\includegraphics[width=1.0\linewidth]{AnonymousSubmission/LaTeX/coop.png}}\\
% \subfloat[CoOp + PLPP]{
% 	\includegraphics[width=1.0\linewidth]{AnonymousSubmission/LaTeX/plpp3.png}}
% % \vspace{10pt}
% \caption{Illustration of (a) CoOp, (b) CoOp + PLPP. In order to integrate perplexity in the training process, our method obtains the labels of vectors by calculating cosine similarity and incorporating a training-free LM Head to output word probability distribution, and then calculate the perplexity of the prompt. Finally, we optimize the prompt together with the cross-entropy loss.}
% % \vspace{10pt}
% \label{fig:plpp}
% \end{figure}

\section{Introduction}
In recent years, the advent of CLIP~\cite{clip} and ALIGN~\cite{align} have driven increased exploration of VL models, which are capable of training and reasoning by using both visual and textual data. It is important to note that such models have a high demand for data and require extensive training on a large-scale image-text pairs to achieve good performance. For instance, the training scheme of CLIP model involves a staggering 400 million image-text pairs. Following the pre-training phase, VL models can perform image classification by employing a carefully crafted prompt, such as ``a photo of a \{category\}," as input for the text encoder. Simultaneously, the image encoder processes the visual input. Subsequently, the classification results are obtained by computing the cosine similarity between text and image representations across all categories.

% \begin{figure}[htb]
% \centering
% %\includegraphics[width=3in]{fig5}
% \subfloat[CoOp]{
% 	\includegraphics[width=1.0\linewidth]{AnonymousSubmission/LaTeX/coop.png}}\\
% \subfloat[CoOp + PLPP]{
% 	\includegraphics[width=1.0\linewidth]{AnonymousSubmission/LaTeX/plpp3.png}}
% % \vspace{10pt}
% \caption{Illustration of (a) CoOp, (b) CoOp + PLPP.}
% % \vspace{10pt}
% \label{fig:plpp}
% \end{figure}

While the development of high-quality contextual prompts~\cite{goodprompt1} has demonstrated the capacity to enhance the performance of CLIP and other similar VL models, it often relies upon a considerable expenditure of time and the specific domain knowledge of human experts. Furthermore, this resource-intensive process may also prove to be ineffective when confronted with novel or unforeseen scenarios. Moveover, the combination of vast parameter space and constraints on available training data, particularly in a few-shot setting, make it infeasible  to fully fine-tune the entire model for downstream tasks.

%Engaging in such fine-tuning carries the added risk of erasing valuable knowledge acquired during the large-scale pretraining phase and introducing the potential for overfitting to the specific downstream task. 
Fine-tuning the entire model on specific tasks risks erasing its general knowledge and can lead to overfitting. Such fine-tuning may hinder the model's adaptability to unforeseen scenarios and reduce its generalization to broader applications.
To address these challenges, inspired by recent advances in Natural Language Processing (NLP)~\cite{attention,nlp1,nlp2,nlp3,nlp4,nlp5,nlp6}, CoOp~\cite{coop} introduces a prompt learning method as an alternative to manually crafting prompts for specific tasks. Diverging from the prior fine-tuning paradigms, CoOp keeps both the image and text encoders of CLIP fixed, exclusively fine-tuning the learnable prompt, which consists of a set of vectors. Following in the footsteps of CoOp, several approaches have been proposed to enhance the training paradigm of prompt or to introduce the learnable prompt to different layers, as exemplified by~\cite{kgcoop,cocoop,maple,proda,prograd,plot,dpt,PromptSRC}. However, these methods still tend to lead models to overfit in downstream tasks.

%\begin{figure}[htbp] 
%	\centering
%	\includegraphics[scale=0.4]{energy_inspired2.png}
%	\caption{Overview of perplexity prediction to two different sentences. Red pot represents the value of perplexity. Low perplexity prediction is assigned to a hand-written sentence, while high perplexity prediction is assigned to a sentence composed of randomly initialized words.}
%	\label{fig:energy}
%\end{figure}
To mitigate the issue of overfitting in prevailing approaches, it is essential to develop new methods that can prevent the model from overtraining. Therefore, we propose a plug-in prompt regularization method called PLPP. The rationale behind our method is straightforward. In the domain of NLP, perplexity is a crucial metric used to evaluate the performance of LMs, where lower perplexity values indicate better model performance in predicting the next word in a sequence of text. In the context of VL models, we fix image and text encoders, integrate perplexity to regularize the process of prompt learning. We use a two-step operation to calculate perplexity. \textbf{a)} obtaining labels: we calculate cosine similarity between the weight of the embedding layer and each vector in prompt, then the index with the largest cosine similarity is used as the corresponding label. \textbf{b)} outputting word probability distribution: We place a LM head, which requires no training, behind the encoder to output the word probability distribution. After these two steps, we can easily calculate the perplexity. 
% In short, perplexity strictly constrains the direction of prompts optimization. It requires the prompts optimization to not only find prompts with good performance in the training task, but also find prompts that are similar to their own outputs after passing through the text encoder and LM head. If the input vector contains the universal features of the input sequence, then the text encoder can better learn the global features of the input sequence, thereby enhancing the generalization ability of the model.
%if we fix the text encoder, the theoretical underpinnings of perplexity can be regarded as preventing the output embeddings of prompt far away from .
%constraining the optimization of prompt within a specific region. 
%\begin{equation} \label{eq:coop}
%%	p(y = i|x) = \frac{exp(sim(g(t_i), f)/\tau)}{\sum_{j=1}^{K}exp(sim(g(t_j), f)/\tau)},
%	\mathbb{E}_{\tilde{s}\sim p_{\theta}(s)}
%\end{equation}

In summary, our main contributions are as follows:
\begin{itemize}
	\item[$\bullet$] We propose PLPP to mitigate the common issue of prompt overfitting in VL models by introducing an explainable metric perplexity. PLPP regularizes prompts optimization by constraining the distribution of prompts close to the output distribution.
	% \\
	\item[$\bullet$] PLPP unites prompt learning and perplexity by incorporating a LM Head that without training. PLPP is a plug-in method, which can be easily integrated into any prompt-based learning methods in VL model without increasing the parameters that need to be optimized.
        \item[$\bullet$] We unveil the essence of PLPP is a self-distillation and turn hard label distribution to soft label distribution to make the model training more stable. We also choose the top-$k$ values in distribution to significantly reduce the computational cost. We introduce mutual self-distillation learning to accelerate model convergence. 
	% \\
	% \item[$\bullet$] We systematically investigate the influence of the hyperparameter of perplexity loss on model performance across various VL tasks. The experiments reveal that employing a moderate coefficient for regularizing prompts can enhance the overall effectiveness of our VL models.
		
\end{itemize} 

% \begin{figure*}[htbp]
% \centering
% %\includegraphics[width=3in]{fig5}
% \subfloat[CoOp]{
% 	\includegraphics[width=0.5\linewidth]{AnonymousSubmission/LaTeX/coop.png}}
% \subfloat[CoOp + PLPP]{
% 	\includegraphics[width=0.5\linewidth]{AnonymousSubmission/LaTeX/plpp3.png}}
% \vspace{10pt}
% \caption{Illustration of (a) CoOp, (b) CoOp + PLPP. In order to integrate perplexity in the training process, our method obtains the labels of vectors by calculating cosine similarity and incorporating a training-free LM Head to output word probability distribution, and then calculate the perplexity of the prompt. Finally, we optimize the prompt together with the cross-entropy loss.}
% \vspace{10pt}
% \label{fig:plpp}
% \end{figure*}
\begin{figure*}[tb]
\centering
\includegraphics[width=1.0\textwidth]{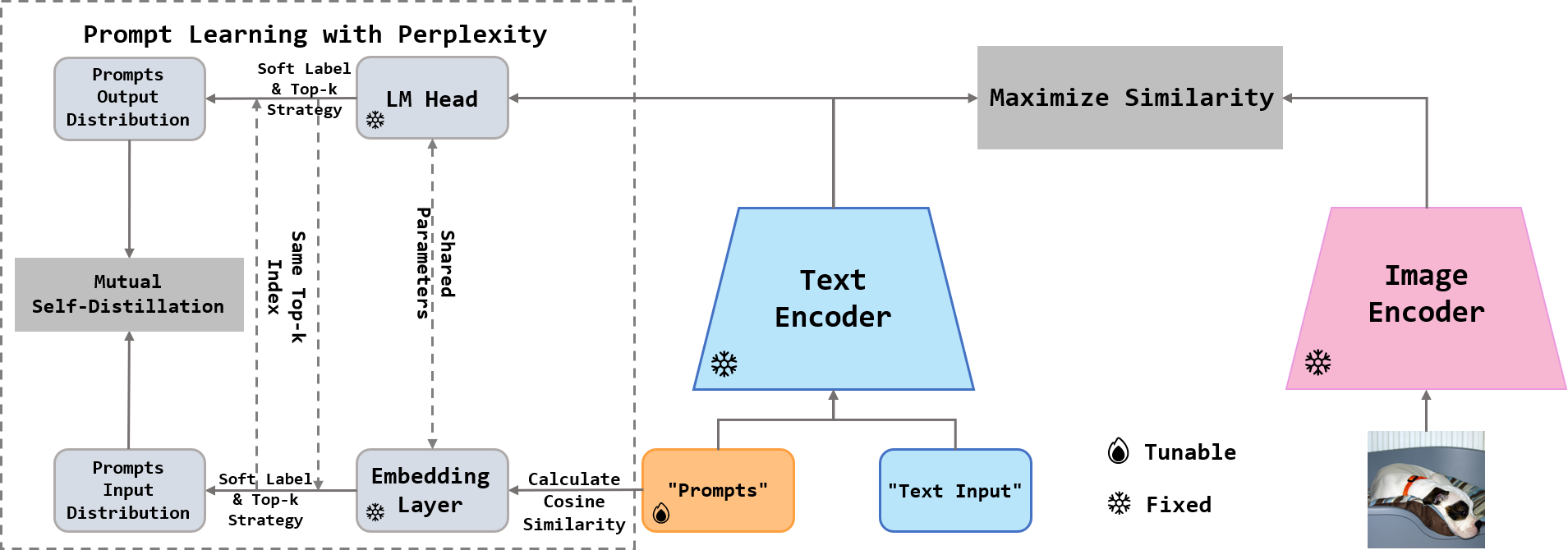}
% \vspace{10pt}
\caption{Overview of our proposed plug-in PLPP (\textbf{P}rompt \textbf{L}earning with \textbf{P}er\textbf{P}lexity) method for prompt learning in VL models.}
% \vspace{10pt}
\label{fig:PLPPSD}
\end{figure*}

\section{Related Works}

\textbf{Pre-training for VL models.} The pre-training phase of the VL model requires unsupervised learning on a large number of image-text pair datasets, due to its voracious appetite for data.. The goal of this phase is to facilitate the model to align the image features with the corresponding text features. CLIP~\cite{clip} and ALIGN~\cite{align} utilize more than four million image-text pairs for pre-training. To ensure the proximity of analogous inputs within the same modality, TCL~\cite{tcl} employs a combination of cross-modal and intra-modal self-supervision, yielding synergistic advantages in representation learning. In a concerted effort to bolster training efficiency, DeCLIP~\cite{declip} not only exploits cross-modal multi-view and intra-modal supervision but also introduces a novel cross-modal Nearest-Neighbor Supervision mechanism, which leverages information emanating from analogous pairs in a more nuanced manner. OneR~\cite{oner} and MS-CLIP~\cite{msclip} adopt a unified transformer encoder architecture for image-text pairs. HiCLIP~\cite{hiclip} enhances the vision and text encoders of CLIP with hierarchy-aware attentions, enabling the model to learn semantic hierarchies in a layer-by-layer fashion.\\
\textbf{Enhancement of Modules in Pre-trained VL Models.} CALIP~\cite{calip} introduces an ingenious attention module devoid of parameters, thereby augmenting the zero-shot performance of CLIP~\cite{clip}. CLIP-Adapter~\cite{clip-adapter}, on the other hand, introduces an additional bottleneck layer into the model. This layer is responsible for acquiring novel features and executing residual-style feature fusion with the originally pretrained features. Meanwhile, Tip-Adapter~\cite{tipadap} inherits the advantageous property of being training-free, as seen in CLIP-Adapter. Furthermore, generating weights through a key-value cache model derived from the few-shot training set enhances the adaptability and effectiveness of the downstream task performance. ATC~\cite{ATC} introduces a novel two-branch architecture. One branch employs ConditionNet to synthesize a textual cache from image features, while the other constructs an learnable visual cache to enhance versatility of the model.\\
% CaFo~\cite{CaFo} leverages four types of prior knowledge from expert models such as CLIP, DINO~\cite{DINO},  DALL-E~\cite{DALL-E}, and GPT-3~\cite{GPT3}, to enrich its limited training data, then aid the few-shot visual recognition tasks. TaskRes~\cite{TaskRes} directly optimize a collection of prior-independent parameters as residual to the features of the text ``a photo of a \{category\}''. ATC~\cite{ATC} introduces a novel two-branch architecture. One branch employs ConditionNet to synthesize a textual cache from image features, while the other constructs an learnable visual cache to enhance versatility of the model.
\textbf{Learnable Prompt for Pre-trained VL Models.} Prompt learning represents significant advancement in the field of NLP. CoOp~\cite{coop} represents a pioneering effort in the field of computer vision to custom extend VL models using this approach. This innovation yields substantial improvements in performance when compared to manually crafted prompts in downstream tasks, particularly in the realm of few-shot classification. Nonetheless, CoOp has limitations in terms of its ability to generalize to broader, unseen categories within the same dataset. In response, CoCoOp~\cite{cocoop} expands this paradigm by training a lightweight neural network to generate an input-conditional token for each image, which leads to enhancing generalization performance. Prompt-Adapter~\cite{prompt-adapter} integrates pre-trained prompt fine-tuning with an efficient adaptation network, achieving enhanced few-shot classification performance. Using prompts in a single branch of CLIP is suboptimal because it limits the model's adaptability to the two representation spaces for adjusting downstream tasks. MaPLe~\cite{maple}, which pioneers prompt learning in both the visual and language branches, has been shown to significantly enhance the alignment of representations. Moreover, MaPLe introduces a profound prompting strategy that extends the scope of prompt learning not only to the input but also across multiple transformer blocks. DPT~\cite{dpt} proposes a novel paradigm that simultaneously incorporates the insights derived from textual and visual prompts. Furthermore, it develops the Class-Aware Visual Prompt Tuning scheme, which generates visual prompts in a dynamic manner based on both task-related and instance-specific prompts. When CoOp-based methodologies are employed in training downstream tasks, learnable prompts tend to accrue task-specific textual knowledge but overlook the essential reservoir of general textual knowledge that underpins robust generalization. KgCoOp~\cite{kgcoop} intervenes to minimize the divergence between the textual embeddings generated by learned prompts and their hand-crafted prompts, averting the loss of essential knowledge. Besides, ProGrad~\cite{prograd} introduces a selective update mechanism for prompts, exclusively attending to those prompts whose gradients align with the gradients of the Kullback-Leibler (KL) loss, calculated by reconciling learnable prompts and hand-crafted prompts. This alignment criterion necessitates that the angle between the two kinds of gradients falls below $90^\circ$. Plot~\cite{plot} uses optimal transport to align text and image output by minimizing transport cost from prompt features to local features of image. Moreover, Plot uses the optimal transport distance to evaluate the match between images and categories, and convert match score to a prediction probability. PromptSRC~\cite{PromptSRC} proposes a self-regularizing framework, which includes mutual agreement maximization, prompt self-ensembling regularization, and textual diversity regularization.

\section{Methodology}
%In this section, we provide an overview of vision-language models about pre-training, inference, and prompt learning. Additionally, we present our proposed solution, Prompt Learning with PerPlexity, which aims to Enhance the comprehensibility of prompt with perplexity while maintain comparable performance in downstream tasks.
In this section, we provide an overview of CoOp, and introduce the evaluation metric perplexity in NLP and what its essence is in the prompt learning of VL models. Additionally, we introduce our method, \textbf{P}rompt \textbf{L}earning with \textbf{P}er\textbf{P}lexity (PLPP), which aims to leverage perplexity to regularize the learning process of prompts, leading to addressing the overfitting issue of prompts.

\subsection{An Overview of CoOp}
%CoOp is the first method that brings prompts learning to pretrained VLMs
CoOp, which is designed to enhance the performance of CLIP in few-shot and domain generalization tasks, introduces prompt learning to VL models. Instead of using hand-crafted prompt templates, CoOp initializes a set of learnable vectors, each vector dimension is 512, which is consistent with the dimension of word embeddings. The number of vectors is usually set to 2, 4, 8, or 16. Concretely, the learnable vector set is denoted as $ \bm{V} = \{\bm{v_1}, \bm{v_2}, \ldots, \bm{v_M}\}$, with $M$ being the count of vectors. Each prompt, denoted as $\bm{p_i} = \{\bm{v_1}, \bm{v_2}, \ldots, \bm{v_M}, \bm{c_i}\}$, amalgamates these learnable vectors with the class token embedding $\bm{c_i}$, where $\bm{c_i}$ represents the tokenized class name corresponding to the $i$-th class. Subsequently, all prompts are feed into CLIP's text encoder, denoted as $g(.)$. Assuming $\bm{f}$ represents the visual embedding of $\bm{x}$, the ultimate prediction probability for predicting the image $x$ as $i$-th class is calculated as follows:
\begin{equation} \label{eq:coop}
	p(y = i|x) = \frac{exp(sim(g(\bm{p_i}), \bm{f})/\tau)}{\sum_{j=1}^{K}exp(sim(g(\bm{p_j}), \bm{f})/\tau)},
\end{equation}
where $sim(., .)$ signifies a metric function such as cosine similarity, and $\tau$ corresponds to the temperature. Finally, given an image and its label, the prediction probability and the label are utilized to compute the cross-entropy loss. During training, only the learnable vectors $\bm{V}$ can be optimized, while the parameters of text and image encoder are frozen.

\subsection{Perplexity}
Perplexity~\cite{mauve} serves as a prominent metric used to assess the quality of a LM, quantifying its capacity to predict a given sequence. At its core, the LM strives to output a probability distribution over a predefined vocabulary of words. Consequently, when subjected to test sets comprising hand-crafted sentences, a higher probability assigned to the corresponding output signifies a superior LM, while conversely, a lower probability suggests otherwise.
For a given sentence in the test set, denoted as $W = \{w_1, w_2, \ldots, w_N\}$, where $N$ signifies the total sentence length, the perplexity of the sentence is calculated as follows:
\begin{equation} \label{eq:ppl}
	\begin{split}
		Perplexity(W) &=P(W)^{-\frac{1}{N}} \\&=  e^{-\frac{1}{N} \sum_{i=1}^{N}\log P(w_i | w_{<i})}
		\\ &= e^{H(Q, P)}.
		%		\sqrt[N]{\frac{1}{p(w_1)p(w_2)...p(w_N)}} \\ &=
	\end{split}
\end{equation}
The calculation of perplexity is as Equation~\ref{eq:ppl}. Here, $P(w_i | w_{<i})$ represents the probability of a word appearing at the $i$-th position in a sentence, with the specific word index being denoted earlier, i.e., model prediction distribution. $Q$ corresponds to the indices of these words in $vocab\_size$ which maps words to integers, i.e. ground truth and $H$ signifies the cross-entropy function. It is worth noting that in the realm of NLP, perplexity is employed to evaluate the efficacy of a LM when exposed to human-written sentences. In the context of our work, we freeze the text encoder, and use perplexity to regularize the learning process of the prompt. Minimizing perplexity is to find prompts that closely match their encoded outputs, thus capturing more global information within each prompt vector. KL divergence can be used to measure the difference in information between the two distributions. In the context of prompt learning for VL models, the KL divergence can be related to perplexity as Equation~\ref{eq:kl} shows:

% \begin{equation} \label{eq:kl}
% 	\begin{split}
% 		KL(Q||P)
% 		&=\sum_{x \in X} Q(x) \log \frac{Q(x)}{P(x)} \\&= \sum_{x \in X} Q(x) \log {Q(x)} - \sum_{x \in X} Q(x) \log P(x)
% 		\\ &= - H(Q)+H(Q, P)=H(Q, P)=\log{PPL}, 
% 		%		\sqrt[N]{\frac{1}{p(w_1)p(w_2)...p(w_N)}} \\ &=
% 	\end{split}
% \end{equation}
\begin{equation} \label{eq:kl}
	\begin{split}
		KL(Q||P)
		&=\sum_{x \in X} Q(x) \log \frac{Q(x)}{P(x)}
		\\ &= - H(Q)+H(Q, P)=H(Q, P)=\log{PPL}, 
		%		\sqrt[N]{\frac{1}{p(w_1)p(w_2)...p(w_N)}} \\ &=
	\end{split}
\end{equation}

where $Q$ and $P$ represent input and output distribution, and $PPL$ is a simplified representation of $Perplexity(W)$ in Equation~\ref{eq:ppl}. Since $Q$ is one-hot distribution, it is obviously $H(Q)=0$ and can be omitted. Thus, perplexity can be regarded as a hard label in self-distillation~\cite{self-distill}, which is a kind of model distillation. Perplexity aligns the input prompt distribution with the output distribution of the text encoder, that is to let shadow layer features learn deep layer features.
\begin{figure*}[htbp]
	\centering
	\subfloat{
		\centering
		\includegraphics[width=0.33\linewidth]{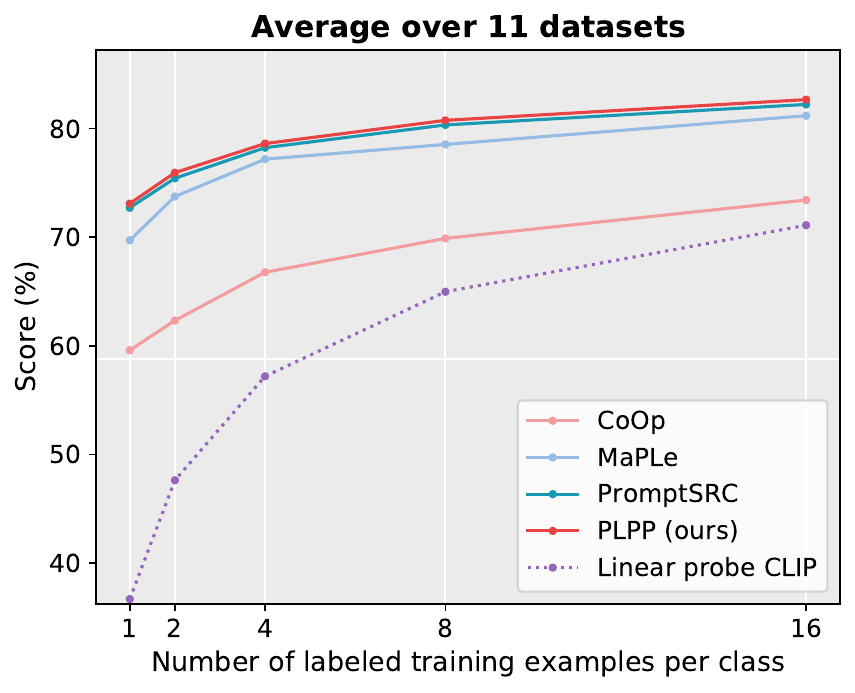}}
	%		\caption{Image 1}
	%		\label{fig:subfig1}
	%	\end{subfig}
% \hfill
\subfloat{
	\centering
	\includegraphics[width=0.33\linewidth]{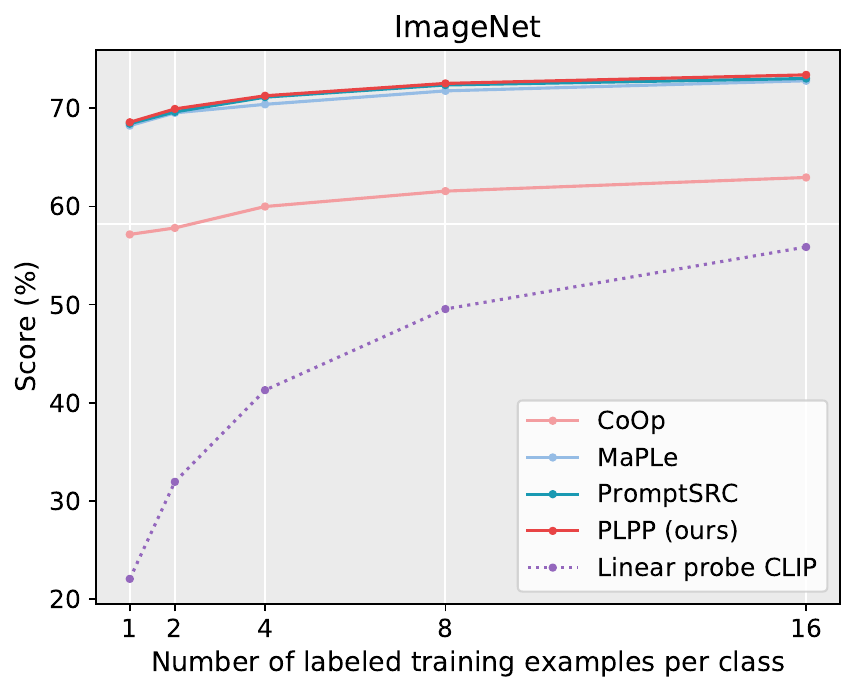}}
%		\caption{Image 2}
%		\label{fig:subfig2}
%	\end{subfig}
% \hfill
%	\begin{subfig}[t]{0.24\linewidth}
\subfloat{
	\centering
	\includegraphics[width=0.33\linewidth]{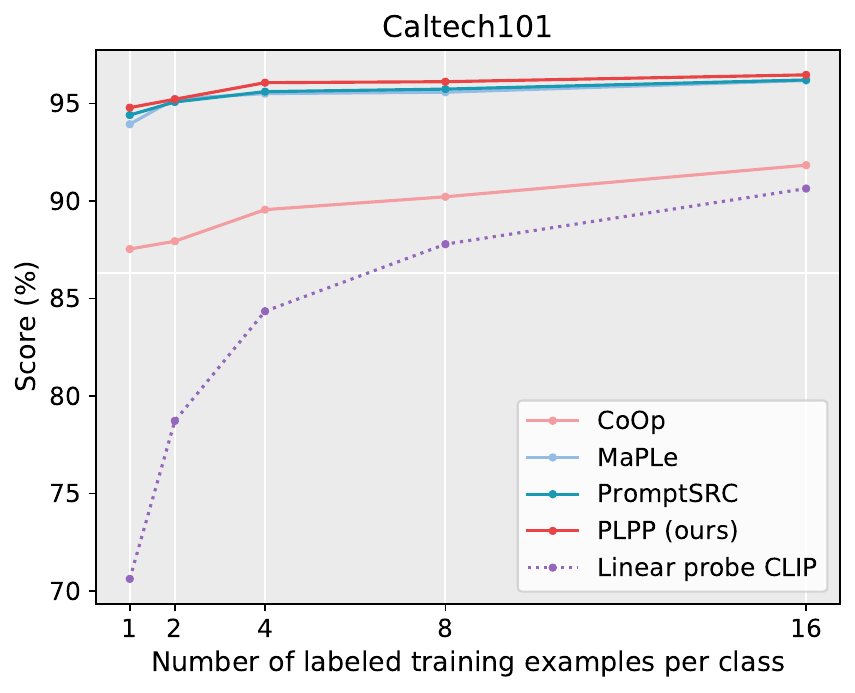}}
%		\caption{Image 1}
%		\label{fig:subfig1}
%	\end{subfig}
%	\hfill

%	\begin{subfig}[t]{0.24\linewidth}
\subfloat{
	\centering
	\includegraphics[width=0.33\linewidth]{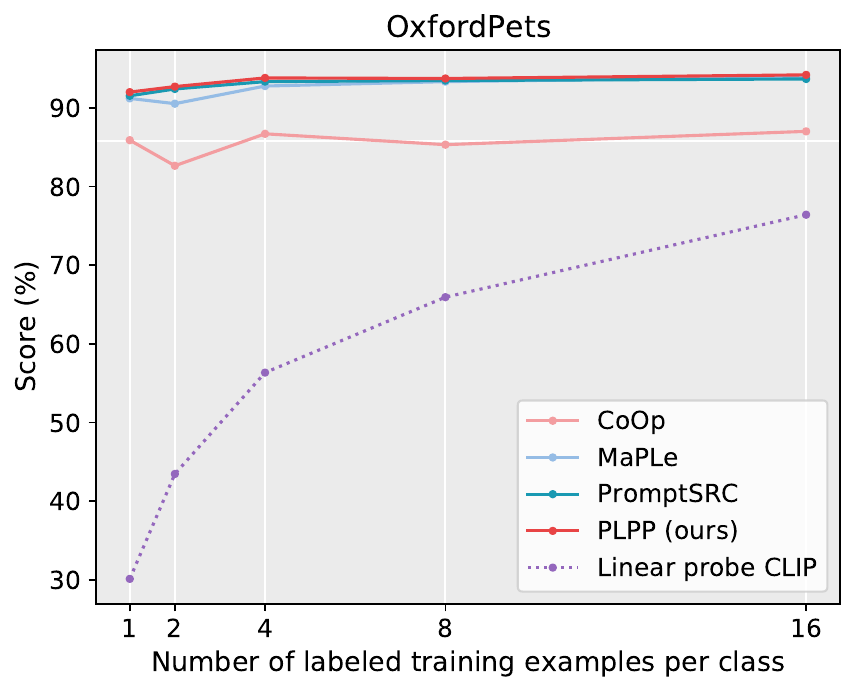}}
%		\caption{Image 2}
%		\label{fig:subfig2}
%	\end{subfig}
% \hfill
%	\begin{subfig}[t]{0.24\linewidth}
\subfloat{
	\centering
	\includegraphics[width=0.33\linewidth]{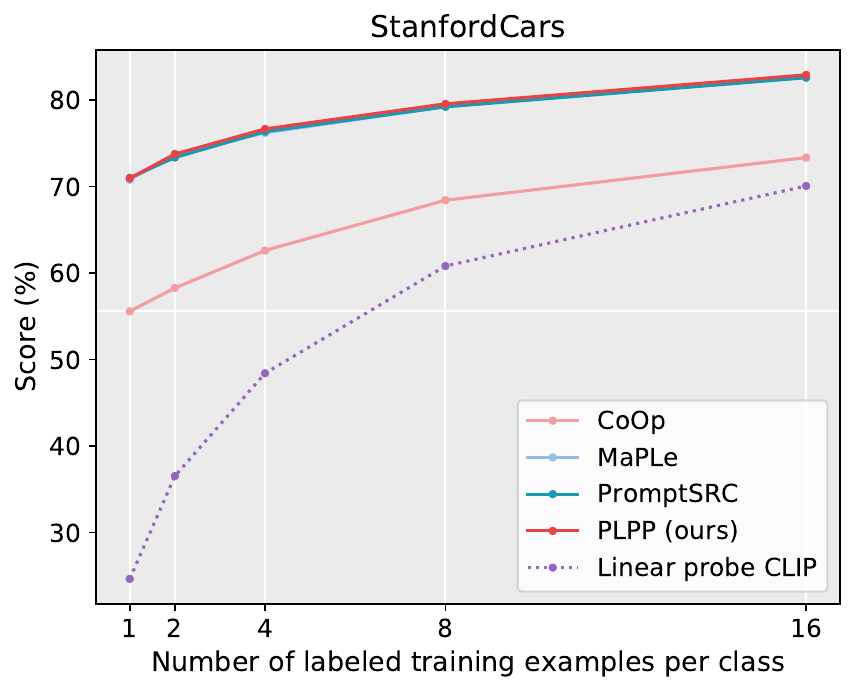}}
%		\caption{Image 1}
%		\label{fig:subfig1}
%	\end{subfig}
% \hfill
%	\begin{subfig}[t]{0.24\linewidth}
\subfloat{
	\centering
	\includegraphics[width=0.33\linewidth]{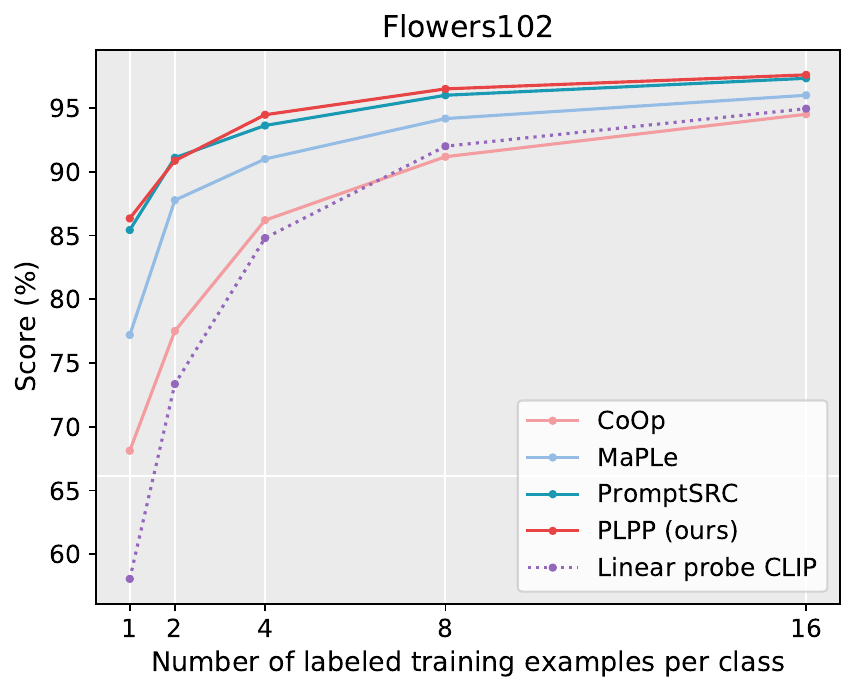}}
%		\caption{Image 2}
%		\label{fig:subfig2}
%	\end{subfig}
%	\hfill

%	\begin{subfig}[t]{0.24\linewidth}
\subfloat{
	\centering
	\includegraphics[width=0.33\linewidth]{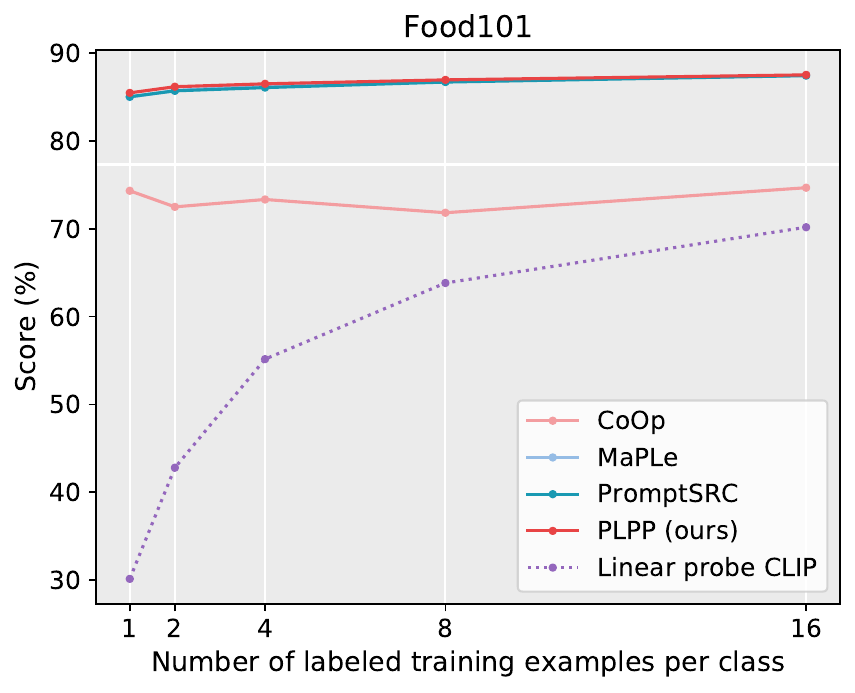}}
%		\caption{Image 1}
%		\label{fig:subfig1}
%	\end{subfig}
% \hfill
%	\begin{subfig}[t]{0.24\linewidth}
\subfloat{
	\centering
	\includegraphics[width=0.33\linewidth]{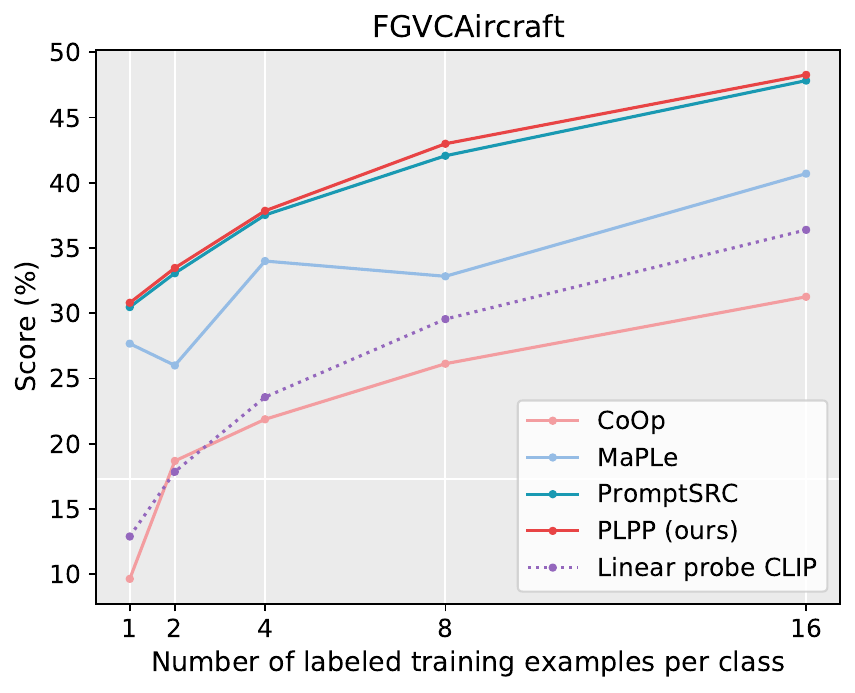}}
%		\caption{Image 2}
%		\label{fig:subfig2}
%	\end{subfig}
% \hfill
%	\begin{subfig}[t]{0.24\linewidth}
\subfloat{
	\centering
	\includegraphics[width=0.33\linewidth]{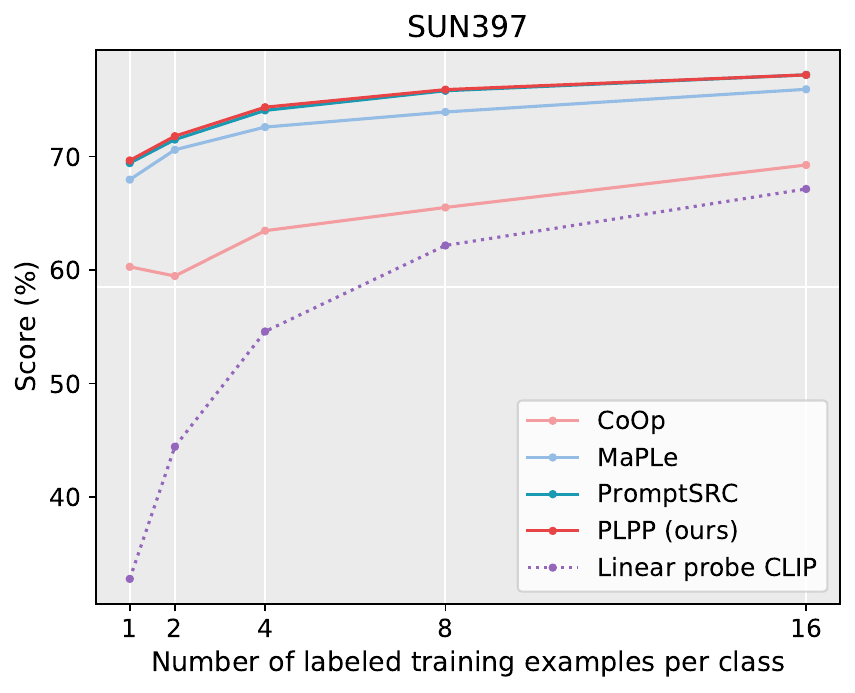}}
%		\caption{Image 2}
%		\label{fig:subfig2}
%	\end{subfig}
% \hfill
%	\begin{subfig}[t]{0.24\linewidth}

\subfloat{
	\centering
	\includegraphics[width=0.33\linewidth]{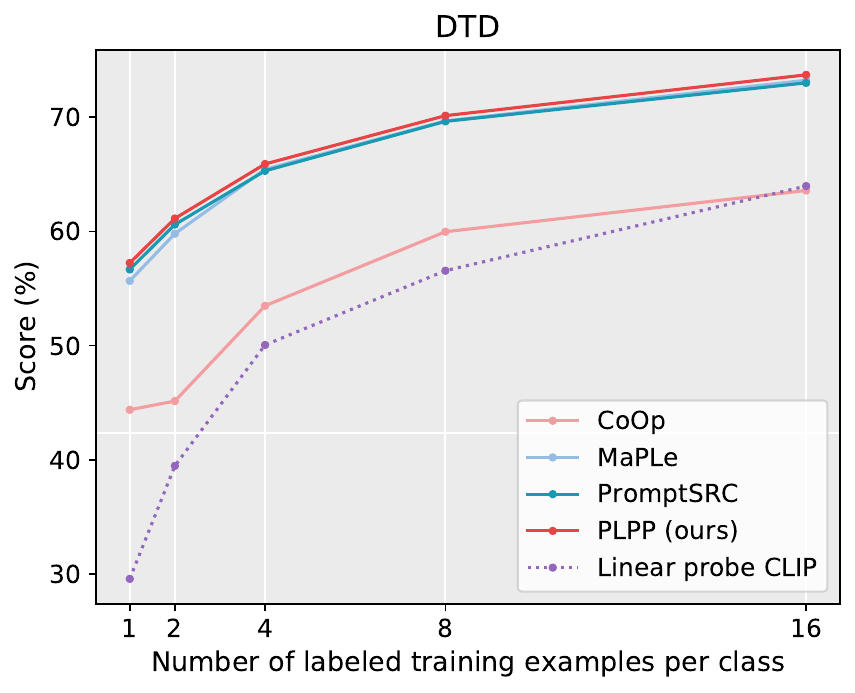}}
%		\caption{Image 2}
%		\label{fig:subfig2}
%	\end{subfig}
% \hfill
%	\begin{subfig}[t]{0.24\linewidth}
\subfloat{
	\centering
	\includegraphics[width=0.33\linewidth]{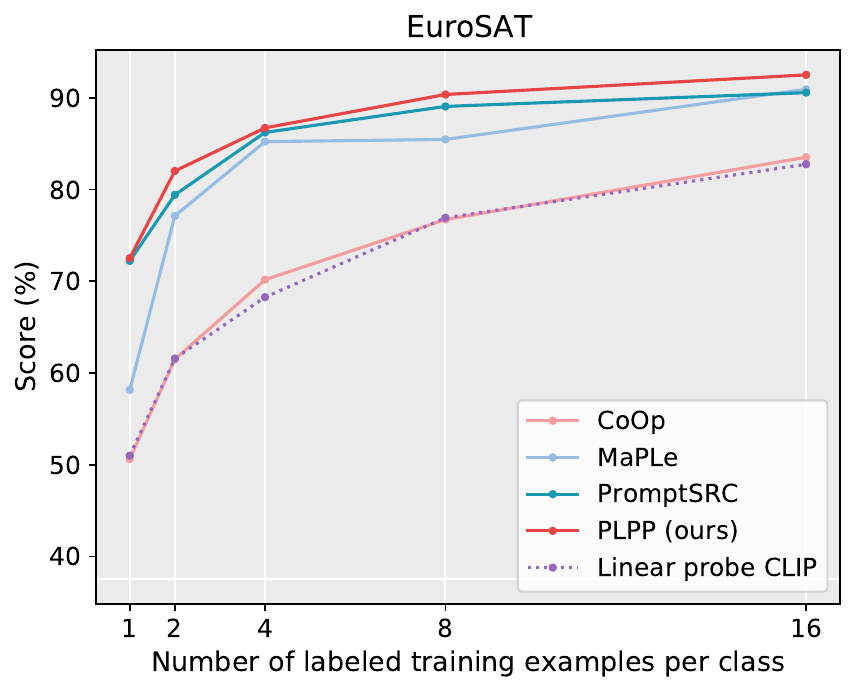}}
%		\caption{Image 2}
%		\label{fig:subfig2}
%	\end{subfig}
% \hfill
%	\begin{subfig}[t]{0.24\linewidth}
\subfloat{
	\centering
	\includegraphics[width=0.33\linewidth]{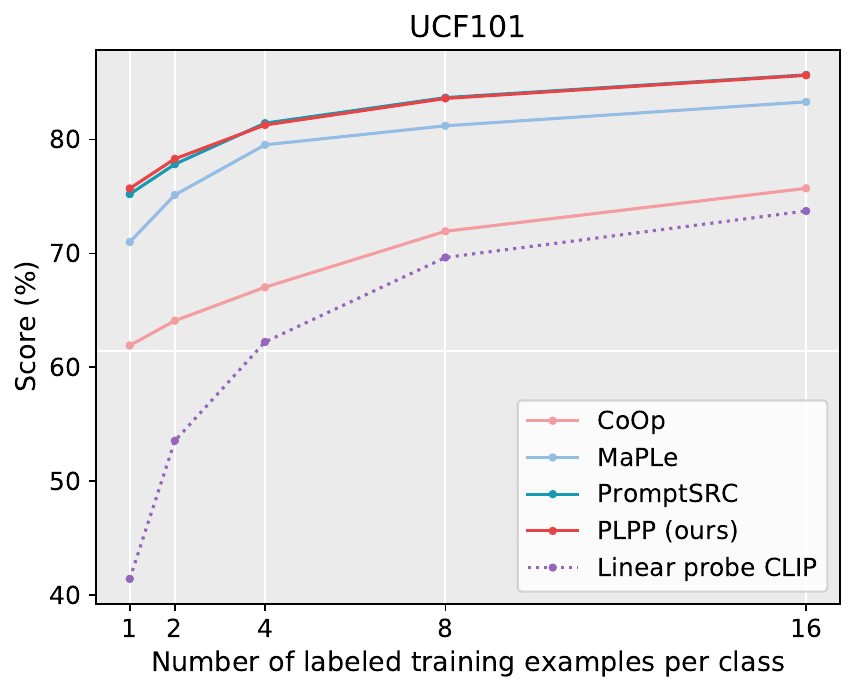}}
%		\caption{Image 2}
%		\label{fig:subfig2}
%	\end{subfig}
%	\hfill
% 其他子图和垂直空间的类似代码
% \vspace{10pt}
\caption{The few-shot classification results on 11 datasets. We compare our PLPP with Liner probe CLIP, CoOp, MaPLe, and PromptSRC, demonstrating consistent and significant performance improvements on most datasets. (The average accuracy on 11 datasets is shown on the left top.)}
\label{fig:subfigure}
%	\label{fig:subfigure}
\end{figure*}

\begin{table*}[tb]
	
	\caption{ Accuracy (\%) for the base-to-novel generalization evaluation. All methods are trained on 16 shots from base classes. H: Harmonic mean. }
 % \fontsize{9pt}{11pt}\selectfont
	% \vspace{10pt}
%	\begin{minipage}{0.3\linewidth}
%		\setlength{\tabcolsep}{2pt} % 减小列之间的间隔
		\centering
		\subfloat[Average over 11 datasets. ]{
            \centering
		\begin{tabular}{lcc|c}
			\toprule
			& \text{Base} & \text{Novel} & $\mathrm{H}$ \\
			\midrule
			% CLIP & 69.34 & 74.22 & 71.70\\
			CoOp & 82.69 & 63.22 & 71.66 \\
			CoCoOp & 80.47 & 71.69 & 75.83 \\
			MaPLe & 82.28 & 75.14 & 78.55 \\
			PromptSRC & 84.26& 76.10 & 79.97 \\
			PLPP (Ours) & \textbf{84.32} & \textbf{76.70} & \textbf{80.33} \\
			\bottomrule
		\end{tabular}}
%	\end{minipage}%
	% \hfill
%	\begin{minipage}{0.3\linewidth}
%		\setlength{\tabcolsep}{2pt} % 减小列之间的间隔
%		\centering
		\subfloat[ImageNet. ]{
            \centering
		\begin{tabular}{lcc|c}
			\toprule
			& \text{Base} & \text{Novel} & $\mathrm{H}$ \\
			\midrule
			% CLIP & 72.43 & 68.14 & 70.22 \\
			CoOp & 76.47 & 67.88 & 71.92 \\
			CoCoOp & 75.98& 70.43 &  73.10 \\
			MaPLe & 76.66 & 70.54 & 73.47 \\
			PromptSRC & 77.60 & 70.73 & 74.01 \\
			PLPP (Ours) & \textbf{77.77} &  \textbf{70.87} & \textbf{74.16} \\
			\bottomrule
		\end{tabular}}
%	\end{minipage}%
	% \hfill
%	\begin{minipage}{0.3\linewidth}
%		\setlength{\tabcolsep}{2pt} % 减小列之间的间隔
%		\centering
		\subfloat[Caltech101. ]{
              \centering
		\begin{tabular}{lcc|c}
			\toprule
			& \text{Base} & \text{Novel} & $\mathrm{H}$ \\
			\midrule
			% CLIP & 96.84 & 94.00 & 95.40 \\
			CoOp & 98.00 & 89.81 & 93.73 \\
			CoCoOp & 97.96  &  93.81 & 95.84 \\
			MaPLe &  97.74 & 94.36 & 96.02 \\
			PromptSRC & 98.10 & 94.03 & 96.02 \\
			PLPP (Ours) & \textbf{98.20} &  \textbf{94.77} & \textbf{96.53} \\
			\bottomrule
		\end{tabular}}
%	  \hspace{\fill}
	\\
%	\end{minipage}
	% \vspace{5pt} % 增加垂直间距
%	\begin{minipage}{0.3\linewidth}
%		\setlength{\tabcolsep}{2pt} % 减小列之间的间隔
%		\centering
%		\subcaption{OxfordPets. }
		\subfloat[OxfordPets. ]{
		\begin{tabular}{lcc|c}
			\toprule
			& \text{Base} & \text{Novel} & $\mathrm{H}$ \\
			\midrule
			% CLIP & 91.17 & 97.26 & 94.12 \\
			CoOp & 93.67 & 95.29 & 94.47 \\
			CoCoOp &  95.20 & 97.69 & 96.43 \\
			MaPLe & 95.43 &\textbf{97.76} & 96.58 \\
			PromptSRC & 95.33 & 97.30 & 96.30 \\
			PLPP (Ours) & \textbf{95.60} &  97.47 & \textbf{96.43} \\
			\bottomrule
		\end{tabular}}
%	\end{minipage}%
	% \hfill
%	\begin{minipage}{0.3\linewidth}
%		\setlength{\tabcolsep}{2pt} % 减小列之间的间隔
%		\centering
%		\subcaption{StanfordCars. }
		\subfloat[StanfordCars.]{
		\begin{tabular}{lcc|c}
			\toprule
			& \text{Base} & \text{Novel} & $\mathrm{H}$ \\
			\midrule
			% CLIP & 63.37 & 74.89 & 68.65 \\
			CoOp & 78.12 & 60.40 & 68.13 \\
			CoCoOp & 70.49 & 73.59 & 72.01 \\
			MaPLe & 72.94 & 74.00& 73.47 \\
			PromptSRC & 78.27 & 74.97 & 76.58 \\
			PLPP (Ours) &\textbf{78.79} &  \textbf{75.43} & \textbf{77.07} \\
			\bottomrule
		\end{tabular}}
%	\end{minipage}%
	% \hfill
%	\begin{minipage}{0.3\linewidth}
%		\setlength{\tabcolsep}{2pt} % 减小列之间的间隔
%		\centering
%		\subcaption{Flowers102. }
		\subfloat[Flowers102.]{
		\begin{tabular}{lcc|c}
			\toprule
			& \text{Base} & \text{Novel} & $\mathrm{H}$ \\
			\midrule
			% CLIP & 72.08 & \textbf{77.80} & 74.83 \\
			CoOp & 97.60 & 59.67 & 74.06 \\
			CoCoOp & 94.87 & 71.75 & 81.71 \\
			MaPLe & 95.92 & 72.46 & 82.56 \\
			PromptSRC & \textbf{98.07} & 76.50 & 85.95 \\
			PLPP (Ours) & 97.85 &  \textbf{77.38} & \textbf{86.42} \\
			\bottomrule
		\end{tabular}}
	\\
%%	\end{minipage}
%%	
	% \vspace{5pt} % 增加垂直间距
%%	
%%	\begin{minipage}{0.3\linewidth}
%%		\setlength{\tabcolsep}{2pt} % 减小列之间的间隔
%%		\centering
%%		\subcaption{Food101. }
		\subfloat[Food101.]{
		\begin{tabular}{lcc|c}
			\toprule
			& \text{Base} & \text{Novel} & $\mathrm{H}$ \\
			\midrule
			% CLIP & 90.10& 91.22 & 90.66 \\
			CoOp & 88.33 & 82.26 & 85.19 \\
			CoCoOp & 90.70 & 91.29& 90.99 \\
			MaPLe & 90.71 & 92.05 & 91.38 \\
			PromptSRC & 90.67 & 91.53 & 91.10 \\
			PLPP (Ours) &  \textbf{90.73} &  \textbf{91.57} & \textbf{91.15} \\
			\bottomrule
		\end{tabular}}
%%	\end{minipage}%
	% \hfill
%%	\begin{minipage}{0.3\linewidth}
%%		\setlength{\tabcolsep}{2pt} % 减小列之间的间隔
%%		\centering
%%		\subcaption{FGVCAircraft. }
		\subfloat[FGVCAircraft.]{
		\begin{tabular}{lcc|c}
			\toprule
			& \text{Base} & \text{Novel} & $\mathrm{H}$ \\
			\midrule
			% CLIP & 27.19& 36.29 & 31.09\\
			CoOp & 40.44 &  22.30 &  28.75 \\
			CoCoOp & 33.41 & 23.71 & 27.74 \\
			MaPLe & 37.44 & 35.61 & 36.50 \\
			PromptSRC & \textbf{42.73} & \textbf{37.87} & \textbf{40.15}\\
			PLPP (Ours) & 42.50 & 37.67 & 39.94 \\
			\bottomrule
		\end{tabular}}
%%	\end{minipage}%
	% \hfill
%%	\begin{minipage}{0.3\linewidth}
%%		\setlength{\tabcolsep}{2pt} % 减小列之间的间隔
%%		\centering
%%		\subcaption{SUN397. }
		\subfloat[SUN397.]{
		\begin{tabular}{lcc|c}
			\toprule
			& \text{Base} & \text{Novel} & $\mathrm{H}$ \\
			\midrule
			% CLIP & 69.36 & 75.35 & 72.23 \\
			CoOp & 80.60 & 65.89 & 72.51 \\
			CoCoOp & 79.74 & 76.86 & 78.27 \\
			MaPLe &  80.82 & 78.70 & 79.75 \\
			PromptSRC & \textbf{82.67} & 78.47 & 80.52 \\
			PLPP (Ours) & 82.43 &  \textbf{78.83} & \textbf{80.59} \\
			\bottomrule
		\end{tabular}}
	\\
%%	\end{minipage}
	% \vspace{5pt} % 增加垂直间距
%%	
%%	\begin{minipage}{0.3\linewidth}
%%		\setlength{\tabcolsep}{2pt} % 减小列之间的间隔
%%		\centering
%%		\subcaption{DTD. }
		\subfloat[DTD.]{
		\begin{tabular}{lcc|c}
			\toprule
			& \text{Base} & \text{Novel} & $\mathrm{H}$ \\
			\midrule
			% CLIP & 53.24 & 59.90& 56.37 \\
			CoOp & 79.44& 41.18 & 54.24 \\
			CoCoOp & 77.01 &  56.00 & 64.85 \\
			MaPLe & 80.36 & 59.18 & 68.16 \\
			PromptSRC & 83.37 & 62.97 & 71.75 \\
			PLPP (Ours)  & \textbf{83.68} & \textbf{63.81}& \textbf{72.41} \\
			\bottomrule
		\end{tabular}}
%%	\end{minipage}%
	% \hfill
%%	\begin{minipage}{0.3\linewidth}
%%		\setlength{\tabcolsep}{2pt} % 减小列之间的间隔
%%		\centering
%%		\subcaption{EuroSAT. }
		\subfloat[EuroSAT.]{
		\begin{tabular}{lcc|c}
			\toprule
			& \text{Base} & \text{Novel} & $\mathrm{H}$ \\
			\midrule
			% CLIP & 56.48& 64.05 & 60.03 \\
			CoOp & 92.19 & 54.74 & 68.69\\
			CoCoOp & 87.49 &  60.04 & 71.21 \\
			MaPLe & \textbf{94.07} & 73.23 & 82.35 \\
			PromptSRC & 92.90 & 73.90 & 82.32 \\
			PLPP (Ours)  & 93.20 & \textbf{76.67} & \textbf{84.13} \\
			\bottomrule
		\end{tabular}}
%	\end{minipage}%
	% \hfill
%	\begin{minipage}{0.3\linewidth}
%		\setlength{\tabcolsep}{2pt} % 减小列之间的间隔
%		\centering
%		\subcaption{UCF101. }
		\subfloat[UCF101.]{
		\begin{tabular}{lcc|c}
			\toprule
			& \text{Base} & \text{Novel} & $\mathrm{H}$ \\
			\midrule
			% CLIP & 70.53 & 77.50 & 73.85 \\
			CoOp & 84.69 & 56.05 & 67.46 \\
			CoCoOp & 82.33 & 73.45& 77.64 \\
			MaPLe & 83.00 & 78.66 & 80.77 \\
			PromptSRC & \textbf{87.10} & 78.80 & 82.74 \\
			PLPP (Ours)  & 86.78 & \textbf{79.19}& \textbf{82.81} \\
			\bottomrule
		\end{tabular}}
%	\end{minipage}
	\label{tab:btng}
\end{table*}

\begin{table*}[tb]
	\centering
	\caption{Comparison of PLPP with existing methods in cross-dataset evaluation setting. All methods are trained on 16 shots.}
	% \vspace{10pt}
	% \resizebox{1.0\textwidth}{!}{
% \small
% {\fontsize{10pt}{1mm}
 % \setlength{\tabcolsep}{1.7mm}{
		\begin{tabular}{lccccccccccccc}
			\toprule
			& \textbf{Source} & &\multicolumn{11}{c}{\textbf{Target}} \\
			\cmidrule{2-2} \cmidrule{4-14} 
			& \rotatebox{90}{ImageNet}  && \rotatebox{90}{Caltech101} & \rotatebox{90}{OxfordPets} & \rotatebox{90}{StanfordCars} & \rotatebox{90}{Flowers102} & \rotatebox{90}{Food101} & \rotatebox{90}{Aircraft} & \rotatebox{90}{SUN397} & \rotatebox{90}{DTD} & \rotatebox{90}{EuroSAT} & \rotatebox{90}{UCF101} & \rotatebox{90}{Average}\\
			\midrule
			\text{CoOp} & \textbf{71.51} && 93.70 & 89.14 & 64.51 & 68.71 & 85.30 & 18.47 & 64.15 & 41.92 & 46.39 & 66.55 &63.88\\
			\text{CoCoOp} & 71.02 && 94.43 & 90.14 & 65.32 & 71.88 & 86.06 & 22.94 & 67.36 & 45.73 & 45.37 & 68.21 &65.74\\
			\text{MaPLe} & 70.72  && 93.53 & \textbf{90.49} & 65.57 & \textbf{72.23} & 86.20 & \textbf{24.74} & 67.01 & 46.49 & \textbf{48.06} & 68.69 &\textbf{66.30}\\
			\text{PromptSRC} & 71.27 && 93.60 & 90.25 & 65.70 & 70.25 & 86.15 & 23.90 & 67.10 & 46.87& 45.50 & 68.75 &65.81\\
			\midrule
			\rowcolor{gray!20}\text{PLPP} & 71.03 && \textbf{94.46} & 90.44 & \textbf{65.87} & 70.89 & \textbf{86.42} & 24.66 & \textbf{67.42}  & \textbf{47.07}	 & 46.83 & \textbf{68.83} &66.29\\
			\bottomrule
			% \\
		\end{tabular}
  % }
		\label{tab:cdg}
	% }
	% \\
\end{table*}
\subsection{Prompt Learning with Perplexity}

In this subsection, we detail the implementation of PLPP, a novel plug-in method that breaks the mold by bridging the gap between perplexity~\cite{mauve} evaluation and prompt learning in VL models. Previously, these two concepts were seen as separate and independent. PLPP unites them in a novel and powerful way. 

From the previous subsection we can know that the relationship between self-distillation and perplexity, which is shown in Equation \ref{eq:kl}. The $Q$ distribution is calculated by using cosine similarity between prompts and embedding layer. As for $P$, we introduce an LM head positioned after the text encoder to output the distribution $P$. “The LM head consists of a simple linear layer without bias, with its weights initialized from the transpose of the $embedding.weight$. Besides, we demonstrate that perplexity can serve as hard label for self-distillation. In order to further prevent overfitting, we replace hard label $Q$ with soft label, and soft label is commonly used in knowledge distillation. Since we introduce the soft label technique and the index is over $40,000$, these increase the computation significantly. To mitigate the computational cost, we employ top-$k$ strategy in $Q$ to retrain only the largest $k$ values, resulting in $topk(Q)$ as the updated $Q$. At the same time, we save the indexes of the largest $k$ values and use these indexes to get the final $P$. To ensure training stability and accelerate model convergence, we introduce mutual self-distillation learning. The perplexity loss is then defined as follows:

\begin{equation} \label{eq:plpp1}
	\mathcal{L}_{PPL} = e^{\frac{1}{2}\cdot KL(Q_1||P_1))} + e^{\frac{1}{2}\cdot KL(P_1||Q_1)},
\end{equation}
\begin{equation} \label{eq:plpp1}
	\mathcal{L}_{IPPL} = e^{\frac{1}{2}\cdot KL(Q_2||P_2))} + e^{\frac{1}{2}\cdot KL(P_2||Q_2)},
\end{equation}

where $Q_1$ is obtained by $topk(Q)$ and the indexes of the largest $k$ values are saved to obtain $P_1$, and $P_2$ is obtained by $topk(P)$ and the indexes of the largest $k$ values are saved to obtain $Q_2$.

Then we calculate the predicted probabilities for all categories on the corresponding dataset according to Equation \ref{eq:coop}. Subsequently, the cross-entropy loss $\mathcal{L}_{CE}$ is based on the prediction probabilities and ground-truth label $y$ for image $x$. The flow chart is shown in Figure~\ref{fig:PLPPSD}. 

% As for perplexity, we introduce an LM head positioned after the text encoder to output word probability distribution $P$. The LM head comprises a straightforward linear layer devoid of bias, with its weight initialized using the transpose of $embedding.weight$, referencing the weight parameter of the embedding layer. Given that perplexity can be expressed in terms of cross-entropy loss and necessitates the labels of prompts, we utilize the dot product to calculate the cosine similarity between $V$ and the weight parameter of the embedding layer. This operation returns the index corresponding to the maximum similarity as the label for the prompts. After each batch, the prompts are updated, the label assignments for the prompts need to be recomputed. The pseudo-code of our PLPP is provided in Appendix.

In summary, we denote $\mathcal{L}_{CE}$ and $\mathcal{L}_{PPL}$, $\mathcal{L}_{IPPL}$ as the loss for aligning image-text features and for regularizing learnable prompts through perplexity. $\alpha$ controls the importance between $\mathcal{L}_{PPL}$ and $\mathcal{L}_{IPPL}$. $\lambda$ controls the weight of the regularization term. We have the overall loss function of PLPP as in Equation~\ref{eq:plpp}.

\begin{equation} \label{eq:plpp}
	\mathcal{L}_{PLPP} = \mathcal{L}_{CE} + \lambda \cdot ( \alpha \cdot\mathcal{L}_{PPL} +(1 - \alpha) \cdot\mathcal{L}_{IPPL} ).
\end{equation}

% \subsection{Theoretical Analysis for PLPP}
% Perplexity loss can be approximated by self-distillation loss $L$ (Eq (3)), representing the KL divergence between the deep output $P_D(x)$ and the shallow output $P_S(x)$:

% \begin{equation} \label{ta1}
%     L=D_{KL}(P_D(x)||P_S(x))=\sum_i P_D(i|x)\log\frac{P_D(i|x)}{P_S(i|x)},
% \end{equation}

% where $x$ is the learnable prompt. In order to minimize $L$, we need to compute the gradient of the input prompt $x$:

% \begin{equation} \label{ta2}
%     \frac{\partial L}{\partial x}=\frac{\partial}{\partial x}\sum_i P_D(i|x)\log \frac{P_D(i|x)}{P_S(i|x)}.
% \end{equation}

% By the chain rule, we can get:

% \begin{align} \label{ta3}
%     \frac{\partial L}{\partial x}&=\frac{\partial}{\partial x}\sum_i P_D(i|x)\log \frac{P_D(i|x)}{P_S(i|x)}\notag\\
%     &=\sum_i\{\frac{\partial P_D(i|x)}{\partial x} \log \frac{P_D(i|x)}{P_S(i|x)}+P_D(i|x) \frac{\partial}{\partial x}\log\frac{P_D(i|x)}{P_S(i|x)}\}.
% \end{align}

% Expanding the second item, we can get:

% \begin{equation} \label{ta4}
%     \frac{\partial}{\partial x}\log \frac{P_D(i|x)}{P_S(i|x)}=\frac{1}{P_D(i|x)}\frac{\partial P_D(i|x)}{\partial x}-\frac{1}{P_S(i|x)} \frac{\partial P_S(i|x)}{\partial x}
% \end{equation}

% Substituting the result into the Equation \ref{ta3}:

% \begin{equation} \label{ta5}
%     \frac{\partial L}{\partial x}=\sum_i \{\frac{\partial P_D(i|x)}{\partial x}(1+\log \frac{P_D(i|x)}{P_S(i|x)})-\frac{P_D(i|x)}{P_S(i|x)}\frac{\partial P_S(i|x)}{\partial x}\}.
% \end{equation}

% As shown in the Equation \label{ta5},

\section{Experiments}
We conduct a comprehensive evaluation of PLPP across four benchmarks for image recognition task: (1) few-shot classification, (2) base-to-novel generalization, (3) cross-dataset evaluation, domain evaluation, and (4) ablation study for the hyperparameter of perplexity loss. PLPP denotes PropmtSRC + PLPP if not otherwise specified in the experiments.

\subsection{Evaluation Settings}
\textbf{Datasets.} In few-shot classification and base-to-novel generalization, we follow to the methodology established by CoOp and CoCoOp, using a total of 11 datasets to evaluate the performance of our method. The 11 datasets include general object recognition datasets: ImageNet~\cite{imagenet} and Caltech101~\cite{caltech101}, fine-grained image recognition datasets: OxfordPets~\cite{oxfordpets}, StanfordCars~\cite{stanfordcars}, Flowers102~\cite{flowers102}, Food101~\cite{food101}, and FGVCAircraft~\cite{fgvc}, a satellite image classification dataset: EuroSAT \cite{eurosat}, an action classification dataset: UCF101 \cite{ucf101}, a texture classification dataset: DTD \cite{dtd}, a scene recognition: SUN397 \cite{sun397}. For domain generalization, ImageNet serves as the source dataset, and the target datasets include ImageNetV2~\cite{imagenetv2}, ImageNet-Sketch~\cite{imagenetsketch}, ImageNet-A~\cite{imagenet-a} and ImageNet-R~\cite{imagenet-r}.
% \\

\textbf{Baselines.} In few-shot classification, we compare PLPP with four baseline methods. The first baseline is Linear probe CLIP, involves training a linear classifier after the CLIP image encoder. CoOp is the second baseline, which learns the unified context prompt through data-driven means instead of time-consuming manual design. The third baseline, MaPLe, is a pioneering work that introduces prompt learning to both the visual and language branches. Lastly, PromptSRC, proposes a self-regularizing method that consists of three components: mutual agreement maximization, prompt self-ensembling regularization, and textual diversity regularization. In all tasks, we use the unified context prompt of CoOp, CoCoOp, MaPLe and PromptSRC as the baseline methods. To further verify the effectiveness of PLPP, we use the best baseline method PromptSRC to integrate our PLPP. For simplicity, we use PLPP to denote PromptSRC + PLPP in all experiments. 
% \\

\textbf{Implementation Details.} In few-shot classification, all the methods are trained with 1, 2, 4, 8, and 16 shots from train set. Subsequently, conducting evaluations on the test dataset. In base-to-novel generalization, cross-dataset evaluation, and domain generalization, all methods are training on 16 shots. To ensure equitable comparisons, we compute the results for all methods and datasets by averaging over three random seeds. For all experiments, we adhere to the guidelines provided in PromptSRC, utilizing vit-b/16~\cite{vit} as the backbone for the image encoder. The number of learnable vectors, denoted as $M$, is consistently set to 4. We adhere to the training epochs, schedule, and data augmentation settings used in PromptSRC. For domain generalization and cross-data evaluation tasks, we set $\lambda$ to 10 and $\alpha$ to 0.2. Since different datasets have different sensitivities to $\mathcal{L}_{PPL}$ and $\mathcal{L}_{IPPL}$, the best $\lambda$ and $\alpha$ varies across different datasets and in few-shot classification and base-to-novel generalization, we list in the appendix the best hyperparameter corresponding to the different datasets. For top-$k$ strategy, we set $k$ to 5 in all tasks.

\subsection{Few-Shot Classification}
\label{sec:fewshot}
%Figure \ref{fig:nineimages} illustrates the comparisons over 11 datasets. Overall, our PLPP achieves clear advantages over baseline models for all few-shot settings on average performance. Specifically, PLPP outperforms CoOp by 9.5\%, 6.9\% and 5.1\% on FGVCAircraft, EuroSAT and Flowers102 given 1 shot, and the average improvement over 11 datasets is 3.2\%. These results demonstrate the anti- overfitting ability of our ProGrad when the samples from downstream tasks are extremely limited. When it comes to 16 shots training, the average improvement induced by ProGrad is less appealing to around 0.5\%. The reason is that the sufficient number of samples from downstream tasks can effectively avoid overfitting. Nonetheless, the average performance gains in all shots settings validate the capability of our ProGrad to improve prompt learning in a data-efficient way.
Figure~\ref{fig:subfigure} demonstrates the comparative analysis across 11 diverse datasets. Overall, our PLPP manifests distinct advantages over the baseline models across various few-shot scenarios, showcasing significant improvement in average performance compared with Linear Probe CLIP, CoOp, and MaPLe. Furthermore, in comparison with PromptSRC, PLPP demonstrates a mean performance lead under 1, 2, 4, 8, and 16 shots across all the datasets. The average improvement in accuracy for each dataset is as follows: OxfordPets (0.4\%), Flowers102 (0.5\%), FGVCAircraft (0.5\%), DTD (0.6\%), EuroSAT (1.3\%), StanfordCars (0.2\%), Food101 (0.3\%), SUN397 (0.2\%), Caltech101 (0.3\%), UCF101 (0.1\%), and ImageNet (0.2\%).

\subsection{Base-to-Novel Generalization}
\label{sec:btng}
%Since CoOp has the weak generalizability problem, which leaves a huge gap between base classes accuracy and unseen classes accuracy, CoCoOp proposes image-conditioned to solve the problem. To compared with CoOp and CoCoOp, our Ex-CoOp use the same network architecture as CoOp, e.g., prompt, text encoder, image encoder.To evaluate the generalization performance from seen classes to unseen classes, we equally divide the classes into two groups, i.e., base classes and new classes. All the methods are only trained on base classes and tested on both base classes and novel classes. We also report the harmonic mean of base-class and new-class accuracies to evaluate the trade-off.
%We compare 
Base-to-novel generalization is that we first partition the dataset into two groups based on the classes, one called base classes and the other called novel classes. Then, we train our model on base classes and evaluate the performance of the model on novel classes. We use a harmonic mean (HM) as a composite measure to evaluate the base versus novel classes performance trade-off, which as calculated by: $H = 2 \cdot\frac{Base \cdot Novel}{Base + Novel}$. As presented in Table~\ref{tab:btng}, our proposed PLPP exhibits the consistent performance advantages in base-to-novel generalization setting on 11 datasets.

PLPP significantly outperforms CoOp, CoCoOp, and MaPLe in both base and novel classes. Against PromptSRC, PLPP achieves better results on 7 out of 11 datasets for base classes and improves on 10 out of 11 datasets for novel classes, with a notable 2.77\% increase on EuroSAT.

\subsection{Cross-Dataset Evaluation}
\label{sec:cdg}
To assess how well PLPP generalizes across different datasets, we train all the methods on ImageNet and test them on 10 other datasets. As shown in Table~\ref{tab:cdg}, PLPP's performance on ImageNet is competitive with other methods. In the 10 target datasets, PLPP exceeds the performance of CoOp and CoCoOp in 10 out of 10 and 9 out of 10 datasets, respectively. PLPP also outperforms MaPLe and PromptSRC in 6 out of 10 and 10 out of 10 datasets. Overall, PLPP performs well compared to PromptSRC. However, its average performance on the target datasets is still lower than MaPLe's, which is mainly due to MaPLe's strong results on the Flowers102 and EuroSAT. The sensitivity of EuroSAT to prompts (due to its only 10 classes) affects performance. Additionally, PromptSRC uses independent vision-language prompt tokens, while MaPLe uses dependent tokens that better capture the relationship between text and images, giving MaPLe an edge in cross-dataset generalization.

\subsection{Domain Generalization}
\label{sec:dg}
%The domain generalization paradigm assesses the models' capacity for generalization in a target domain distinct from the source domain. Traditional fine-tuning with a limited dataset from a specific domain can potentially mislead the model into acquiring spurious correlations or patterns confined to that domain, thus yielding a biased model that exhibits subpar performance in unfamiliar domains [1, 28, 36]. In contrast, zero-shot CLIP does not exploit such spurious correlations or patterns since it refrains from fine-tuning on that distribution [36]. Given that our PLPP leverages general knowledge from the pre-trained domain to constrain the fine-tuning process for a particular distribution, our PLPP demonstrates resilience in the face of distributional shifts. As exemplified in Table 2, our PLPP unequivocally surpasses CoOp across all target datasets and even outperforms CoCoOp in three out of four target datasets. It is worth noting that CoCoOp achieves commendable performance in domain generalization through dynamic instance-conditional prompts. However, the computation of instance-conditional prompts significantly escalates training time. Conversely, our PLPP employs a static prompt to mitigate training costs while still outperforming CoCoOp (refer to Table 5).
Domain generalization evaluates a model's ability to generalize to a target domain that is different but related to the source domain.
We train our model on ImageNet dataset, and evaluate it on four specially designed benchmark datasets.
% Traditional fine-tuning on limited data from a specific domain can inadvertently lead the model to acquire spurious correlations.This results in a biased model with poor performance on the target domains. 
As Table~\ref{tab:dge} shows, PLPP consistently outperforms all competing methods across all target datasets with an overall highest average accuracy of 60.8\%. Especially for ImageNet-Sketch, PLPP leads propmtsrc by 0.3\%. The results demonstrate that our PLPP method can get better generalization for datasets with domain shifts.

\begin{table}[tbp]
	\centering
	\caption{Comparison of PLPP with existing methods in domain generalization setting. All methods are trained on 16 shots.}
	% \vspace{10pt}
	% \resizebox{1.0\linewidth}{!}{
 % {\fontsize{10pt}{1mm}
         \setlength{\tabcolsep}{1mm}{
		\begin{tabular}{lccccccc}
			\toprule
			& \textbf{Source} & &\multicolumn{5}{c}{\textbf{Target}} \\
			\cmidrule{2-2} \cmidrule{4-8} 
			& \text{ImageNet}  && \text{-V2} & \text{-S} & \text{-A} & \text{-R} & \text{Avg.}\\
			\midrule
			% \text{CLIP} & 66.7&& 60.8 & 46.2 & 47.8 & 74.0 & 57.2\\
			\text{CoOp} & \textbf{71.5} && 64.2 & 48.0 & 49.7 & 75.2& 59.3 \\
			\text{CoCoOp} & 71.0  && 64.1 & 48.8 & 50.6 & 76.2 & 59.9\\
			\text{MaPLe} & 70.7 && 64.1 & 49.2 & 50.9 & 77.0 & 60.3\\
			\text{PromptSRC} & 71.3 && 64.4 & 49.6 & 50.9 & 77.8 &60.7 \\
			\midrule
			\rowcolor{gray!20}\text{PLPP} & 71.0 && \textbf{64.6} & \textbf{49.7} & \textbf{51.2} & \textbf{77.8} & \textbf{60.8}\\
			\bottomrule
		\end{tabular}
		\label{tab:dge}
  }
	% }
	%	\label{tab:dge}
\end{table}

\section{Conclusion}
%In this paper, we pointed out the over-fitting issues of existing prompt tuning methods for few-shot generalization, which heavily relies on early stopping and data augmentation to promote zero- shot inference. We proposed a prompt tuning method ProGrad that regularize each tuning step not to conflict with the general knowledge of the hand-crafted prompt. Experiments on few-shot classification, base-to-new generalization and domain generalization over 11 datasets demonstrate the effectiveness and efficiency of our PLPP. In the future, we will explore a more powerful metric and intergrate it in the training process of prompts to reach the original intention: the prompts like a sentence by people and high performance.
Prompt-based learning methods greatly reduce the number of parameters that need to be optimized when the VL model is fine-tuned on downstream tasks, and have achieved astonishing performance on various downstream tasks. However, existing prompt learning methods still ignore the problem that prompts are prone to overfitting, thereby damaging the inherent generalization ability of VL models. Our work proposes a plug-in prompt-regularization learning method called PLPP, which addresses the prompt overfitting problem for better generalization. We reveal the essence of perplexity in prompt learning for VL models is a form of self-distillation. To calculate the perplexity loss, we use embedding layer to obtain label and introduce a LM head that requires no training to output word distribution. Moreover, we introduce soft label and top-$k$ strategy to further prevent overfitting and reduce the computational cost. We also employ mutual self-distillation learning to accelerate model convergence. Extensive experiments on four classification tasks show the effectiveness of our PLPP.

\bibliography{aaai25}

\begin{thebibliography}{47}
\providecommand{\natexlab}[1]{#1}

\bibitem[{Bossard, Guillaumin, and Van~Gool(2014)}]{food101}
Bossard, L.; Guillaumin, M.; and Van~Gool, L. 2014.
\newblock Food-101--mining discriminative components with random forests.
\newblock In \emph{ECCV}, 446--461.

\bibitem[{Chen et~al.(2023)Chen, Yao, Song, Li, Rao, and Zhang}]{plot}
Chen, G.; Yao, W.; Song, X.; Li, X.; Rao, Y.; and Zhang, K. 2023.
\newblock Prompt Learning with Optimal Transport for Vision-Language Models.
\newblock In \emph{ICLR}, 1--13.

\bibitem[{Cimpoi et~al.(2014)Cimpoi, Maji, Kokkinos, Mohamed, and Vedaldi}]{dtd}
Cimpoi, M.; Maji, S.; Kokkinos, I.; Mohamed, S.; and Vedaldi, A. 2014.
\newblock Describing textures in the wild.
\newblock In \emph{CVPR}, 3606--3613.

\bibitem[{Deng et~al.(2009)Deng, Dong, Socher, Li, Li, and Fei-Fei}]{imagenet}
Deng, J.; Dong, W.; Socher, R.; Li, L.-J.; Li, K.; and Fei-Fei, L. 2009.
\newblock Imagenet: A large-scale hierarchical image database.
\newblock In \emph{CVPR}, 248--255.

\bibitem[{Dosovitskiy et~al.(2021)Dosovitskiy, Beyer, Kolesnikov, Weissenborn, Zhai, Unterthiner, Dehghani, Minderer, Heigold, Gelly et~al.}]{vit}
Dosovitskiy, A.; Beyer, L.; Kolesnikov, A.; Weissenborn, D.; Zhai, X.; Unterthiner, T.; Dehghani, M.; Minderer, M.; Heigold, G.; Gelly, S.; et~al. 2021.
\newblock An image is worth 16x16 words: Transformers for image recognition at scale.
\newblock In \emph{ICLR}, 1--13.

\bibitem[{Fei-Fei, Fergus, and Perona(2007)}]{caltech101}
Fei-Fei, L.; Fergus, R.; and Perona, P. 2007.
\newblock Learning generative visual models from few training examples: An incremental bayesian approach tested on 101 object categories.
\newblock In \emph{CVIU}, 59--70.

\bibitem[{Gao et~al.(2024)Gao, Geng, Zhang, Ma, Fang, Zhang, Li, and Qiao}]{clip-adapter}
Gao, P.; Geng, S.; Zhang, R.; Ma, T.; Fang, R.; Zhang, Y.; Li, H.; and Qiao, Y. 2024.
\newblock Clip-adapter: Better vision-language models with feature adapters.
\newblock In \emph{IJCV}, 581--595.

\bibitem[{Gao, Fisch, and Chen(2021)}]{nlp1}
Gao, T.; Fisch, A.; and Chen, D. 2021.
\newblock Making Pre-trained Language Models Better Few-shot Learners.
\newblock In \emph{ACL}, 3816--3830.

\bibitem[{Geng et~al.(2023)Geng, Yuan, Tian, Chen, and Zhang}]{hiclip}
Geng, S.; Yuan, J.; Tian, Y.; Chen, Y.; and Zhang, Y. 2023.
\newblock HiCLIP: Contrastive language-image pretraining with hierarchy-aware attention.
\newblock In \emph{ICLR}, 1--13.

\bibitem[{Guo et~al.(2023)Guo, Zhang, Qiu, Ma, Miao, He, and Cui}]{calip}
Guo, Z.; Zhang, R.; Qiu, L.; Ma, X.; Miao, X.; He, X.; and Cui, B. 2023.
\newblock Calip: Zero-shot enhancement of clip with parameter-free attention.
\newblock In \emph{AAAI}, 746--754.

\bibitem[{Hantao~Yao(2023)}]{kgcoop}
Hantao~Yao, C.~X., Rui~Zhang. 2023.
\newblock Visual-Language Prompt Tuning with Knowledge-guided Context Optimization.
\newblock In \emph{CVPR}, 6757--6767.

\bibitem[{Helber et~al.(2018)Helber, Bischke, Dengel, and Borth}]{eurosat}
Helber, P.; Bischke, B.; Dengel, A.; and Borth, D. 2018.
\newblock Eurosat: A novel dataset and deep learning benchmark for land use and land cover classification.
\newblock In \emph{IGARSS}, 204--207.

\bibitem[{Hendrycks et~al.(2021{\natexlab{a}})Hendrycks, Basart, Mu, Kadavath, Wang, Dorundo, Desai, Zhu, Parajuli, Guo et~al.}]{imagenet-r}
Hendrycks, D.; Basart, S.; Mu, N.; Kadavath, S.; Wang, F.; Dorundo, E.; Desai, R.; Zhu, T.; Parajuli, S.; Guo, M.; et~al. 2021{\natexlab{a}}.
\newblock The many faces of robustness: A critical analysis of out-of-distribution generalization.
\newblock In \emph{ICCV}, 8320--8329.

\bibitem[{Hendrycks et~al.(2021{\natexlab{b}})Hendrycks, Zhao, Basart, Steinhardt, and Song}]{imagenet-a}
Hendrycks, D.; Zhao, K.; Basart, S.; Steinhardt, J.; and Song, D. 2021{\natexlab{b}}.
\newblock Natural adversarial examples.
\newblock In \emph{CVPR}, 15262--15271.

\bibitem[{Jang et~al.(2023)Jang, Kong, Jeon, Kim, and Kwak}]{oner}
Jang, J.; Kong, C.; Jeon, D.; Kim, S.; and Kwak, N. 2023.
\newblock Unifying Vision-Language Representation Space with Single-Tower Transformer.
\newblock In \emph{AAAI}, 980--988.

\bibitem[{Jia et~al.(2021)Jia, Yang, Xia, Chen, Parekh, Pham, Le, Sung, Li, and Duerig}]{align}
Jia, C.; Yang, Y.; Xia, Y.; Chen, Y.-T.; Parekh, Z.; Pham, H.; Le, Q.; Sung, Y.-H.; Li, Z.; and Duerig, T. 2021.
\newblock Scaling up visual and vision-language representation learning with noisy text supervision.
\newblock In \emph{ICML}, 4904--4916.

\bibitem[{Jiang et~al.(2020)Jiang, Xu, Araki, and Neubig}]{nlp2}
Jiang, Z.; Xu, F.~F.; Araki, J.; and Neubig, G. 2020.
\newblock How can we know what language models know?
\newblock In \emph{TACL}, 423--438.

\bibitem[{Jin et~al.(2022)Jin, Cheng, Shen, Chen, and Ren}]{goodprompt1}
Jin, W.; Cheng, Y.; Shen, Y.; Chen, W.; and Ren, X. 2022.
\newblock A good prompt is worth millions of parameters: Low-resource prompt-based learning for vision-language models.
\newblock In \emph{ACL}, 2763--2775.

\bibitem[{khattak et~al.(2023)khattak, Rasheed, Maaz, Khan, and Khan}]{maple}
khattak, M.~U.; Rasheed, H.; Maaz, M.; Khan, S.; and Khan, F.~S. 2023.
\newblock MaPLe: Multi-modal Prompt Learning.
\newblock In \emph{CVPR}, 19113--19122.

\bibitem[{Khattak et~al.(2023)Khattak, Wasim, Naseer, Khan, Yang, and Khan}]{PromptSRC}
Khattak, M.~U.; Wasim, S.~T.; Naseer, M.; Khan, S.; Yang, M.-H.; and Khan, F.~S. 2023.
\newblock Self-regulating prompts: Foundational model adaptation without forgetting.
\newblock In \emph{ICCV}, 15190--15200.

\bibitem[{Krause et~al.(2013)Krause, Stark, Deng, and Fei-Fei}]{stanfordcars}
Krause, J.; Stark, M.; Deng, J.; and Fei-Fei, L. 2013.
\newblock 3d object representations for fine-grained categorization.
\newblock In \emph{ICCVW}, 554--561.

\bibitem[{Lester, Al-Rfou, and Constant(2021)}]{nlp3}
Lester, B.; Al-Rfou, R.; and Constant, N. 2021.
\newblock The Power of Scale for Parameter-Efficient Prompt Tuning.
\newblock In \emph{EMNLP}, 3045--3059.

\bibitem[{Li and Liang(2021)}]{nlp4}
Li, X.~L.; and Liang, P. 2021.
\newblock Prefix-tuning: Optimizing continuous prompts for generation.
\newblock In \emph{ACL}, 4582--4597.

\bibitem[{Li et~al.(2022)Li, Liang, Zhao, Cui, Ouyang, Shao, Yu, and Yan}]{declip}
Li, Y.; Liang, F.; Zhao, L.; Cui, Y.; Ouyang, W.; Shao, J.; Yu, F.; and Yan, J. 2022.
\newblock Supervision Exists Everywhere: A Data Efficient Contrastive Language-Image Pre-training Paradigm.
\newblock In \emph{ICLR}, 1--13.

\bibitem[{Lu et~al.(2022)Lu, Liu, Zhang, Liu, and Tian}]{proda}
Lu, Y.; Liu, J.; Zhang, Y.; Liu, Y.; and Tian, X. 2022.
\newblock Prompt distribution learning.
\newblock In \emph{CVPR}, 5206--5215.

\bibitem[{Maji et~al.(2013)Maji, Rahtu, Kannala, Blaschko, and Vedaldi}]{fgvc}
Maji, S.; Rahtu, E.; Kannala, J.; Blaschko, M.; and Vedaldi, A. 2013.
\newblock Fine-grained visual classification of aircraft.
\newblock \emph{arXiv preprint arXiv:1306.5151}.

\bibitem[{Nilsback and Zisserman(2008)}]{flowers102}
Nilsback, M.-E.; and Zisserman, A. 2008.
\newblock Automated flower classification over a large number of classes.
\newblock In \emph{ICVGIP}, 722--729.

\bibitem[{Parkhi et~al.(2012)Parkhi, Vedaldi, Zisserman, and Jawahar}]{oxfordpets}
Parkhi, O.~M.; Vedaldi, A.; Zisserman, A.; and Jawahar, C. 2012.
\newblock Cats and dogs.
\newblock In \emph{CVPR}, 3498--3505.

\bibitem[{Pillutla et~al.(2021)Pillutla, Swayamdipta, Zellers, Thickstun, Welleck, Choi, and Harchaoui}]{mauve}
Pillutla, K.; Swayamdipta, S.; Zellers, R.; Thickstun, J.; Welleck, S.; Choi, Y.; and Harchaoui, Z. 2021.
\newblock MAUVE: Measuring the Gap Between Neural Text and Human Text using Divergence Frontiers.
\newblock In \emph{NeurIPS}, 4816--4828.

\bibitem[{Radford et~al.(2021)Radford, Kim, Hallacy, Ramesh, Goh, Agarwal, Sastry, Askell, Mishkin, Clark et~al.}]{clip}
Radford, A.; Kim, J.~W.; Hallacy, C.; Ramesh, A.; Goh, G.; Agarwal, S.; Sastry, G.; Askell, A.; Mishkin, P.; Clark, J.; et~al. 2021.
\newblock Learning transferable visual models from natural language supervision.
\newblock In \emph{ICML}, 8748--8763.

\bibitem[{Recht et~al.(2019)Recht, Roelofs, Schmidt, and Shankar}]{imagenetv2}
Recht, B.; Roelofs, R.; Schmidt, L.; and Shankar, V. 2019.
\newblock Do imagenet classifiers generalize to imagenet?
\newblock In \emph{ICML}, 5389--5400.

\bibitem[{Shin et~al.(2020)Shin, Razeghi, Logan~IV, Wallace, and Singh}]{nlp5}
Shin, T.; Razeghi, Y.; Logan~IV, R.~L.; Wallace, E.; and Singh, S. 2020.
\newblock Autoprompt: Eliciting knowledge from language models with automatically generated prompts.
\newblock In \emph{EMNLP}, 4222--4235.

\bibitem[{Soomro, Zamir, and Shah(2012)}]{ucf101}
Soomro, K.; Zamir, A.~R.; and Shah, M. 2012.
\newblock UCF101: A dataset of 101 human actions classes from videos in the wild.
\newblock \emph{arXiv preprint arXiv:1212.0402}.

\bibitem[{Sun et~al.(2023)Sun, Qin, Lin, and Chen}]{prompt-adapter}
Sun, J.; Qin, J.; Lin, Z.; and Chen, C. 2023.
\newblock Prompt Tuning based Adapter for Vision-Language Model Adaption.
\newblock \emph{arXiv preprint arXiv:2303.15234}.

\bibitem[{Vaswani et~al.(2017)Vaswani, Shazeer, Parmar, Uszkoreit, Jones, Gomez, and Kaiser}]{attention}
Vaswani, A.; Shazeer, N.; Parmar, N.; Uszkoreit, J.; Jones, L.; Gomez, A.~N.; and Kaiser. 2017.
\newblock Attention Is All You Need.
\newblock In \emph{NeurIPS}, 5998--6008.

\bibitem[{Wang et~al.(2019)Wang, Ge, Lipton, and Xing}]{imagenetsketch}
Wang, H.; Ge, S.; Lipton, Z.; and Xing, E.~P. 2019.
\newblock Learning robust global representations by penalizing local predictive power.
\newblock In \emph{NeurIPS}, 10506--10518.

\bibitem[{Xiao et~al.(2010)Xiao, Hays, Ehinger, Oliva, and Torralba}]{sun397}
Xiao, J.; Hays, J.; Ehinger, K.~A.; Oliva, A.; and Torralba, A. 2010.
\newblock SUN database: Large-scale scene recognition from abbey to zoo.
\newblock In \emph{CVPR}, 3485--3492.

\bibitem[{Xing et~al.(2023)Xing, Wu, Cheng, Zhang, Liang, Wang, and Zhang.}]{dpt}
Xing, Y.; Wu, Q.; Cheng, D.; Zhang, S.; Liang, G.; Wang, P.; and Zhang., Y. 2023.
\newblock Dual Modality Prompt Tuning for Vision-Language Pre-Trained Model.
\newblock In \emph{TMM}, 1--13.

\bibitem[{Yang et~al.(2023)Yang, Meng, Chen, Liu, and Zhang}]{ATC}
Yang, C.; Meng, F.; Chen, S.; Liu, M.; and Zhang, R. 2023.
\newblock Instance-Wise Adaptive Tuning and Caching for Vision-Language Models.
\newblock In \emph{ECAI}, 2834--2841.

\bibitem[{Yang et~al.(2022)Yang, Duan, Tran, Xu, Chanda, Chen, Zeng, Chilimbi, and Huang}]{tcl}
Yang, J.; Duan, J.; Tran, S.; Xu, Y.; Chanda, S.; Chen, L.; Zeng, B.; Chilimbi, T.; and Huang, J. 2022.
\newblock Vision-Language Pre-Training with Triple Contrastive Learning.
\newblock In \emph{CVPR}, 15671--15680.

\bibitem[{You et~al.(2022)You, Zhou, Xiao, Codella, Cheng, Xu, Chang, and Yuan}]{msclip}
You, H.; Zhou, L.; Xiao, B.; Codella, N.; Cheng, Y.; Xu, R.; Chang, S.-F.; and Yuan, L. 2022.
\newblock Learning visual representation from modality-shared contrastive language-image pre-training.
\newblock In \emph{ECCV}, 69--87.

\bibitem[{Zhang et~al.(2019)Zhang, Song, Gao, Chen, Bao, and Ma}]{self-distill}
Zhang, L.; Song, J.; Gao, A.; Chen, J.; Bao, C.; and Ma, K. 2019.
\newblock Be your own teacher: Improve the performance of convolutional neural networks via self distillation.
\newblock In \emph{ICCV}, 3713--3722.

\bibitem[{Zhang et~al.(2022)Zhang, Zhang, Fang, Gao, Li, Dai, Qiao, and Li}]{tipadap}
Zhang, R.; Zhang, W.; Fang, R.; Gao, P.; Li, K.; Dai, J.; Qiao, Y.; and Li, H. 2022.
\newblock Tip-adapter: Training-free adaption of clip for few-shot classification.
\newblock In \emph{ECCV}, 493--510.

\bibitem[{Zhong, Friedman, and Chen(2021)}]{nlp6}
Zhong, Z.; Friedman, D.; and Chen, D. 2021.
\newblock Factual probing is [mask]: Learning vs. learning to recall.
\newblock In \emph{NAACL-HLT}, 5017--5033.

\bibitem[{Zhou et~al.(2022{\natexlab{a}})Zhou, Yang, Loy, and Liu}]{cocoop}
Zhou, K.; Yang, J.; Loy, C.~C.; and Liu, Z. 2022{\natexlab{a}}.
\newblock Conditional Prompt Learning for Vision-Language Models.
\newblock In \emph{CVPR}, 16795--16804.

\bibitem[{Zhou et~al.(2022{\natexlab{b}})Zhou, Yang, Loy, and Liu}]{coop}
Zhou, K.; Yang, J.; Loy, C.~C.; and Liu, Z. 2022{\natexlab{b}}.
\newblock Learning to Prompt for Vision-Language Models.
\newblock In \emph{IJCV}, 2337--2348.

\bibitem[{Zhu et~al.(2023)Zhu, Niu, Han, Wu, and Zhang}]{prograd}
Zhu, B.; Niu, Y.; Han, Y.; Wu, Y.; and Zhang, H. 2023.
\newblock Prompt-aligned Gradient for Prompt Tuning.
\newblock In \emph{ICCV}, 15613--15623.

\end{thebibliography}

\end{document}